\documentclass[10pt,twocolumn,letterpaper]{article}

\usepackage{cvpr}
\usepackage{times}
\usepackage{epsfig}
\usepackage{graphicx}
\usepackage{amsmath}
\usepackage{amssymb}
\usepackage{booktabs}
\usepackage{multirow}
\usepackage{xcolor}
\usepackage{algpseudocode}
\usepackage{algorithm}
\usepackage{bm}
\usepackage{amsfonts}

\newcommand{\RR}{\mathcal{R}}

\newcommand{\EE}{\mathbb{E}}
\newcommand{\DD}{\mathbb{D}}
\newcommand{\TT}{\mathbb{T}}
\newcommand{\II}{\mathcal{I}}
\newcommand{\VV}{\mathcal{V}}

\usepackage[pagebackref=false,breaklinks=true,letterpaper=true,colorlinks,bookmarks=false]{hyperref}

\cvprfinalcopy 

\def\cvprPaperID{****} 

\ifcvprfinal\fi
\begin{document}

\title{AVD: Adversarial Video Distillation}

\author{Mohammad Tavakolian$^1$, Mohammad Sabokrou$^2$, and Abdenour Hadid$^1$ \vspace{5pt}\\ 
 $^1$Center for Machine Vision and Signal Analysis (CMVS), University of Oulu \\
 $^2$Institute for Research in Fundamental Sciences (IPM)
}

\maketitle

\begin{abstract}
In this paper, we present a simple yet efficient approach for video representation, called Adversarial Video Distillation (AVD).
 The key idea is to represent videos by compressing them in the form of realistic images, which can be used in a variety of video-based scene analysis applications. Representing  a video  as  a single image enables us to address the problem of video analysis by image analysis techniques.
To this end, we exploit a 3D convolutional encoder-decoder network to encode the input video  as an image by minimizing the reconstruction error. Furthermore, weak supervision  by an adversarial training procedure is imposed on the output of the encoder to generate semantically realistic images. 
 The encoder learns to extract semantically meaningful representations from a given input video by mapping the 3D input into a 2D latent representation. The obtained representation can be simply used as the input of deep models pre-trained on images for video classification. We evaluated the effectiveness of our proposed method for video-based activity recognition on three standard and challenging benchmark datasets, \ie UCF101, HMDB51, and Kinetics. The experimental results demonstrate that AVD achieves interesting performance, outperforming the state-of-the-art methods for video classification.
\end{abstract}

\section{Introduction}

\label{sec:introduction}

Humans perceive their surrounding environment through the visual information. The human's visual system can efficiently combine the appearance and dynamic information of a scene to comprehend complex environments. On the contrary to human visual system, recognizing and understanding complex scenes is a challenging task for computers. The several approaches, which have addressed the related tasks to video analysis (such as object recognition~\cite{redmon2016you}, scene understanding~\cite{cordts2016cityscapes}, and semantic segmentation~\cite{long2015fully}), indicate that videos provide more information than images. Consequently, focusing on video-based approaches instead of image-based ones, is a reasonable way for achieving better performance for complex tasks such as scene understanding. This gives more importance to investigate video analysis approaches. Evidently, obtaining efficient and discriminative representations goes a long way in video understanding~\cite{Bilen2016}. However, such representation is still a subject of contention and it is not clear what an optimal representation is.
\begin{figure}[t!]
\begin{center}
\addtolength{\tabcolsep}{-5pt}    
\begin{tabular}{ccccccc}
{\footnotesize $\mathcal{V}:$} & \raisebox{-.5\height}{\includegraphics[width=0.27\linewidth]{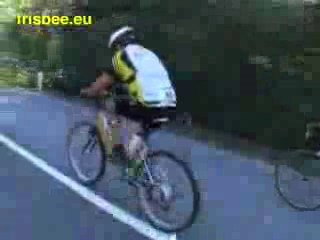}} &
\raisebox{-.5\height}{\includegraphics[width=0.27\linewidth]{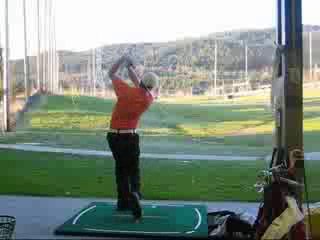}} &
\raisebox{-.5\height}{\includegraphics[width=0.27\linewidth]{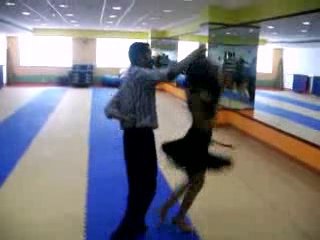}} &~~ &   \\
{\footnotesize  AVD $(\mathcal{V})$:} & \raisebox{-.5\height}{\includegraphics[width=0.27\linewidth]{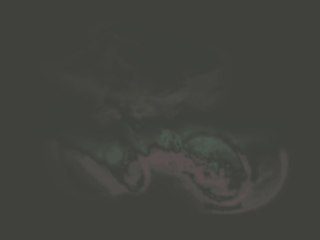}} &
\raisebox{-.5\height}{\includegraphics[width=0.27\linewidth]{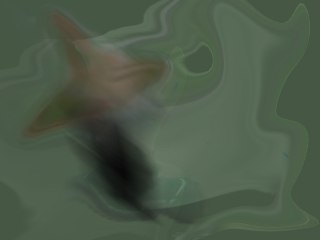}} &
\raisebox{-.5\height}{\includegraphics[width=0.27\linewidth]{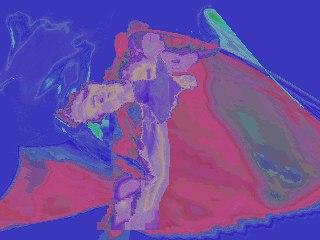}}  \\
\end{tabular}
\addtolength{\tabcolsep}{5pt}    
\end{center}
   \caption{Examples    of   videos  (denoted by $\mathcal{V}$) and their representation using AVD.  AVD  represents both  spatial and temporal characteristics of raw videos  as an RGB image (\ie discriminative feature map)  which can be used as the input of deep models pre-trained on still images.}
 \label{fig:Demo}
 \vspace{-15pt}
\end{figure}

Intuitively, due to a number of challenges such as illumination variations, viewpoint changes, and  camera motions achieving discriminative representations of videos is a non-trivial task. These challenges drastically degrade the performance of video analysis methods. In the past years, a substantial number of approaches have been introduced to cope with these challenges~\cite{Tavakolian2018, tavakolian2019spatiotemporal, cordts2016cityscapes, Wang2018, Tran2015}. Preliminary works treated videos as either sequences of still images or volumetric objects, and applied handcrafted local descriptors on a stack of images~\cite{Wang2013, Willems2008}. The advent of representation learning has introduced methods that learn complex underlying structures in videos. Convolutional Neural Networks (CNNs) convincingly have demonstrated superior performance of representation learning methods~\cite{Krizhevsky2012}. Thanks to large-scale training datasets~\cite{Deng2009, Zhou2014}, CNNs have rapidly taken over the majority of still image-based recognition tasks such as object, scene, and face recognition~\cite{Zhou2014,tavakolian2019spatiotemporal,Taigman2014}. CNNs have also been developed for video representation. However, videos are characterized by the temporal evolution of appearance governed by the motion. It is noteworthy  that due to the simplicity of the structure of images compared to videos, CNNs have shown a better performance in image representation than in video analysis.

Recurrent Neural Network (RNN) have been widely used for learning the representations of sequential data like videos~\cite{Simonyan2014,Tran2015}. However, videos can be of arbitrary lengths and extending deep architecture to yet another dimension of complexity is a non-trivial task as the number of parameters grows significantly. Moreover, it is well-substantiated that efficient representation of appearance and dynamics of a scene is of paramount importance in modeling of videos~\cite{Bilen2016}. In addition, the application of the majority of existing deep models for video representation is limited to trimmed videos. On the contrary, a good model should be able to deal with untrimmed video data in the real-world setting, where events may only occur in a small portion of the entire length of the video. In summary,  the main difficulties and challenges for efficient video representation learning  are: (1) complexity of the video's structure compared to the image and, (2) uncertainty on the duration of the video.  

To overcome these challenges, we propose a novel method called Adversarial Video Distillation (AVD) for video representation, which encodes the appearance and dynamics of videos into a distilled image representation. Following the same goal as~\cite{Bilen2016}, our proposed AVD is based on a completely different and simpler approach. AVD is learned to compress the spatio-temporal information of a video in the format of an RGB image. This image can be fed to any existing deep models devised for still images. AVD uses a deep 3D convolutional encoder network to map the input video into a 2D latent representation. Furthermore, the encoder network adversarially competes with a discriminator network to reconstruct the encoded video from the latent space, \ie RGB image.  In other words, the encoder tries to  represent the  videos via supervision of another network that knows the notion of real scenes. Among the existing video representation methods, \cite{Bilen2016} is closest to ours. However, our AVD has some advantages over~\cite{Bilen2016}: (1)  AVD achieves representations that uniquely characterize the scene, while dynamic image, which is a pooling-based method, imposes equal importance on all frame that might be unfavorable; (2) our method follows a straightforward unsupervised end-to-end process. This provides the opportunity of using a plethora of unlabeled data for training the model.

Our proposed method relaxes the need for a large volume of training data, since the pre-trained still image-based models are deployed for video classification. To provide better intuitions, Figure~\ref{fig:Demo} illustrates some visualizations of the encoded still images from videos using  AVD.

The main contributions of this paper are: (1) proposing an end-to-end deep neural network to distill the spatio-temporal information of videos as an image representation, (2) following an unsupervised training procedure, we train the model in an adversarial manner to learn a semantic image representation from the input video, and (3) achieving state-of-the-art results for video classification on three benchmark datasets, \ie UCF101~\cite{Soomro2012}, HMDB51~\cite{Kuehne2011}, and Kinetics~\cite{Kay2017}. 

\section{Related Work}
\label{sec:related_work}

In the early stages, video representation methods treated videos as a sequence of still images or as a smooth evolution of consecutive frames. By considering the video as a stack of still frames, several spatiotemporal feature extraction methods have been proposed~\cite{Wang2013,Willems2008,Wang2011}. Those methods define a local spatiotemporal neighborhood around each point of interest and a histogram descriptor is extract to capture the spatial and temporal information. Then, some aggregation approaches generate a holistic representation from the local descriptors. Although these handcrafted features are effective for video representation, they lose the discriminative capacity in the presence of camera motion and some other variations and distortions. 

To overcome the limitations of conventional methods for video representation, several works have attempted to learn visual representations by using Convolutional Neural Networks (CNNs). To capture the appearance and dynamics of the video, CNNs have been extended to temporal domain by adding another dimension. Tran \etal~\cite{Tran2015} studied 3D CNN~\cite{Ji2013} on realistic (captured in the wild) and large-scale video databases. Their C3D model learns both the spatial and temporal information in a short segment of the video using 3D convolution operations. Carreira \etal~\cite{Carreira2017} proposed a two-stream inflated 3D CNN (I3D) by converting vanilla Inception-V1 architecture~\cite{Ioffe2015} to a 3D model. They replaced 2D kernels of Inception-V1 to 3D kernel in which the model can use the knowledge of pre-trained 2D model on ImageNet database~\cite{Russakovsky2015}. Qiu \etal~\cite{Qiu2017} developed a Pseudo-3D Residual Network by applying a spatiotemporal factorization on a residual learning module. Diba \etal~\cite{Diba2018} embedded a temporal transition layer in the DenseNet architecture~\cite{Huang2017} and replaced 2D convolutional filters and pooling layers with their 3D counterparts. Despite the fact that 3D CNN-based architectures perform reasonably well in capturing spatiotemporal information, they usually need a lot of training data to achieve a good representation of the video due to their huge number of parameters.

The aforementioned methods only capture local spatiotemporal information within a small time window. Hence, they are not capable of capturing long-range dynamics. Recently, Want \etal~\cite{Wang2018} proposed a temporal segment network to model long-range temporal structure of actions within a video. The authors randomly selected snippets of the video and extracted optical flow and RGB differences from frames that are fed to CNN models for feature extraction. Their method achieves a global representation of the video using a segmental consensus function to aggregate the information from different snippets of the video. Bilen \etal~\cite{Bilen2016} introduced dynamic image by employing a rank pooling technique to capture the temporal evolution of actions and representing the video as one RGB image. They distill the appearance and dynamics of a scene into one single image, which is fed 2D CNN models for action classification. However, pooling techniques consolidate data into compact representations. These techniques also impose equal importance on all frames, which is not favorable. Wang \etal~\cite{Wang2018a} proposed SVM pooling for video summarization. They reformulated the pooling problem as a multiple instance learning context and learned useful decision boundaries on the frame level features from each video against background features.

\section{Adversarial Video Distillation}
\label{sec:method}
\begin{figure*}[t]
    \centering
    \includegraphics[width=0.8\linewidth]{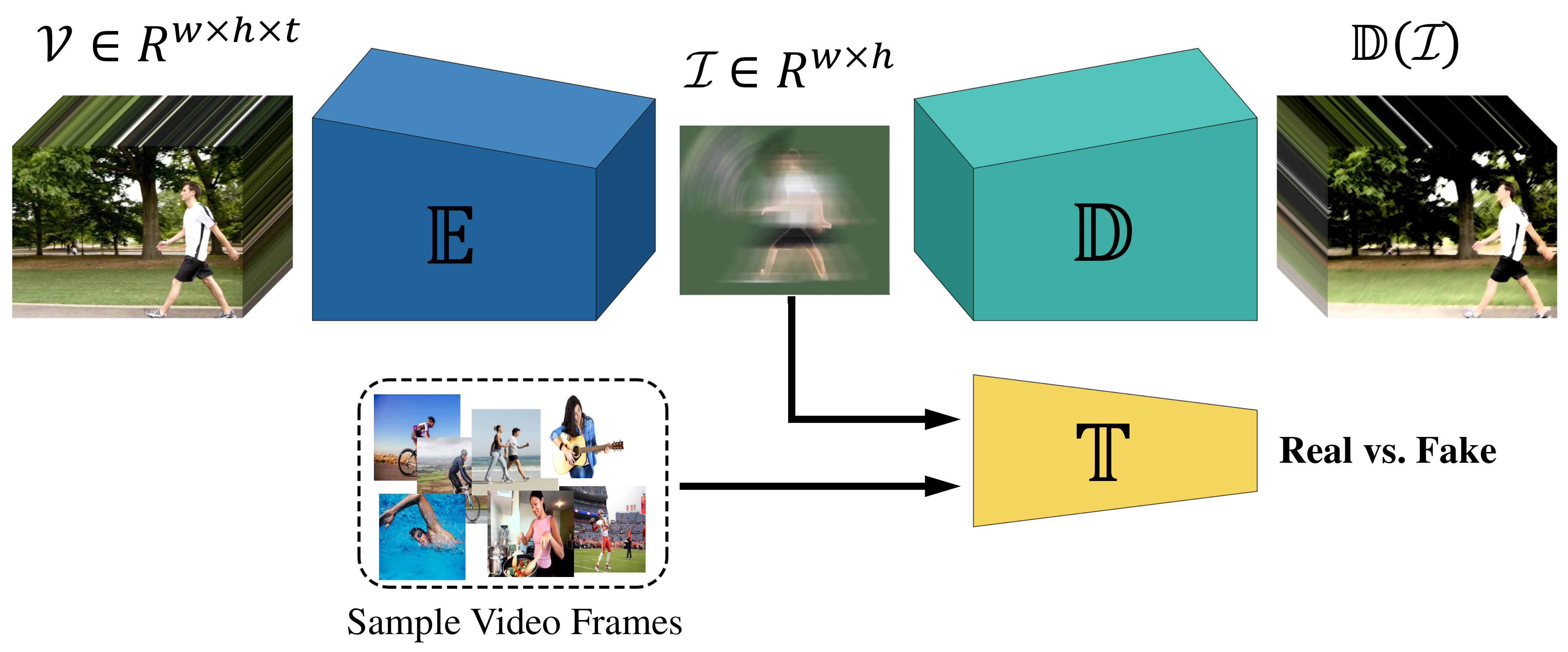}
    \caption{The outline of our proposed AVD for video representation. The encoder network $\EE$ encodes the input video $\VV$ into a distilled image representation $\II$ that can be used as the input of pre-trained deep models.}
    \label{fig:model}
    \vspace{-15pt}
\end{figure*}

Our proposed method, \ie AVD, is an end-to-end deep neural network, which learns to represent videos as single images in an unsupervised adversarial manner. The proposed method is composed of three main networks: $\EE$,  $\DD$, and $\TT$. $\EE$ is an encoder network ($\EE$), and  learns to compute  informative representations of input videos, \ie $\VV \in \RR^{w \times h \times t}$ in form of an image $\II \in \RR^{w \times h}$.  To guarantee that important features of the video are preserved in its representation, \ie $\II$, the encoder is trained jointly with a decoder ($\DD$) network to minimize a reconstruction error. Howbeit by optimizing the  $\EE+\DD$, the $\EE$ is able to distill the video information and compress it into the single image format, the obtained representations are not always meaningful and similar to the  realistic images. To address this issue, the $\EE$ network is supervised by a teacher network ($\TT$). $\TT$  accesses to a large number of real images and knows the distribution of realistic images. Figure~\ref{fig:model} shows the outline of the proposed AVD.  The details of $\EE$, $\TT$ and  $\DD$ networks and also the training procedure of the AVD are explained in the following subsections.      

\subsection{$\EE$ncoder Network: Video Distillation}
Previously,  different versions of encoder-decoder networks have been widely used for unsupervised feature learning \cite{Tavakolian2018}.  After training  these networks, the encoder can map the input data into a latent representation. Inspired by the substantial achievements of auto-encoders, we present an efficient network ($\EE$) to encode the input videos. $\EE$ encodes the appearance and motion of the input video $\VV$ as a distilled image representation containing the gist of the scene. The main difference between $\EE$ and the conventional encoder networks, which are devised for video representation include: (1) the $\EE(\VV)$ is forced to generate a meaningful representation of the input video, (2) by imposing an extra constraint on the output of $\EE$, the $\EE(\VV)$ has to follow the distribution of the video frames. To meet these two properties, $\EE$ should be trained jointly with two deep networks ($\DD$ and $\TT$). In other words, $\EE$ is responsible for distillation of the information, and $\TT$ and $\DD$ are only used for training of $\EE$.

The encoder network $\EE$ is devised based on the 3D CNN. It maps the input video into a RGB image representation by spatiotemporal down-sampling. As an encoder network, $\EE$ is trained to summarize the motion information of the video sequence preserving the appearance information to achieve a discriminative representation of the scene. We call this process video distillation since it captures the spatiotemporal information of the video and encodes it into an unique image representation.

\subsection{$\DD$ecoder Network: Video Reconstruction}
Learning  a decoder  network jointly with an encoder network for data compression and representation by  encoder is a commonplace approach. Similarly, we learn $\EE$ and $\DD$  in terms of an encoder-decoder network.   Successful  reconstruction of the video $\VV$  by $\DD$ from its representation,  \ie $\EE(\VV)$,  means that  $\EE(\VV)$ do not discard the vital information of the video. As $\EE$ must preserve the spatial-temporal information of video, it is optimized to  minimize the  reconstruction errors \ie $e = ||\DD(\II)-\VV||^2$. 

The $\DD$ network is a 3D CNN that up-samples the the latent representation $\EE(\VV)$ to reconstruct the input video by minimizing the reconstruction error. This means the output of $\DD$ should be as close as to the input of the model. Hence, the training of the encoder-decoder network can be performed in an unsupervised end-to-end manner using stochastic gradient descent. The $\DD$ basically mirrors the encoder network $\EE$ in terms of the architecture, except that there are no pooling or linear layers in our network. The rationale behind following this approach is exploring the underlying complex structure of the video.

\subsection{$\TT$eacher Network}
By learning the above-mentioned networks, $\EE$ is able to encode the videos as 2D image representations, which might not exhibit any semantic relationship with the input video. In other words, the obtained image representations may not be good representatives for the input videos provided that there is not any explicit supervision on the training of the auto-encoder. To tackle this problem, we take advantage of the GANs idea to force $\EE$ to generate meaningful representations from the realistic data. So, we add a teacher network $\TT$ that is responsible for supervision of the training of $\EE$ by continuously comparing the output of $\EE$ with the distribution of the video frames in an adversarial manner.  In this way, the $\TT$ network teaches latent space of $\EE+\DD$ to be similar to realistic images. In context of GAN, $\EE$ and $\TT$  can be considered as a generator and a discriminator networks, respectively. Hence, the entire of our proposed model is trained in an unsupervised adversarial fashion.

The $\TT$ network can be defined as a binary classifier for distinguishing between generated samples by $\EE$ and the realistic images. Therefore, the encoder ensures the aggregated posterior distribution can fool the discrimintive adversarial network by thinking that the latent representation $\mathcal{I}$ comes from the true prior distribution of the real video frames. The architecture of $\TT$ is devised based on a stack of multiple fully connected layers to build a binary classifier.

\subsection{$\EE+\DD+\TT$ Learning}
 AVD is trained end-to-end. It includes three networks: $\EE$, $\DD$ and $\TT$. These networks collaborate with each other in the training phase, but only $\EE$  is responsible for video distillation. In other words, two other networks assist this network to successfully learn distilled image representations. The $\EE$ network is trained by the received feedback from $\DD$  and $\TT$ to generate a discriminative representation that capture the appearance and motion information. During the training process. $\EE$ encodes a batch of training videos $V={\VV_i}, i=1\dots, N$, where $N$ is the batch size. Then, $\DD$ reconstructs the videos by mapping the $\EE(V)$ to the original video space. The $\DD$ attempts to reconstruct the input video with  the lowest error. Consequently, the $\EE+\DD$, as an encoder-decoder network, is trained by optimizing the reconstruction errors \ie  $\mathcal{L}_{\EE+\DD}$:
\begin{equation}
   \mathcal{L}_{\EE+\DD}=\sum_{i=1}^N{||\VV_i-\DD(\EE(\VV_i))||^2}
\end{equation}
  
Concurrently with $\mathcal{L}_{\EE+\DD}$ optimization,  the $\TT$ checks whether the latent representation of $\EE+\DD$ network, \ie the output of $\EE$, follows the distribution of the video frames or not. $\TT$ has access to a large number of video frames. Hence, it knows the distribution of the different scenes denoted by $p_r$.  In this framework, $\EE$ is assumed a network that tries to fool the $\TT$ by convincing it that its output is an real image, not a generated  ones. These two networks are learning in an adversarial manner based on the GAN theory. Once the training process is over, the encoder network is able to generate meaningful image representations of the input videos. Similar to GANs, $\EE+\DD$ network is learned as a min-max game and by optimizing the following objective: 
\begin{multline}
\mathcal{L}_{\EE+\TT}= \min_\EE \max_\TT \Big( {E}_{\II \sim  p_r}[\log(\mathcal{\TT}(\II))] + \\ {E}_{\VV \sim  p_v}[\log(1-\TT(\underbrace{\EE(\VV)}_{\II}] \Big) 
\label{eq:q2}
\end{multline} 
where $p_\VV$ is the distribution of realistic videos. The AVD needs to train based on both   $\mathcal{L}_{\EE+\DD}$ and $\mathcal{L}_{\EE+\TT}$ objective functions. So, the objective function of AVD is defined as: 
\begin{equation}
\begin{aligned}
\centering
    \mathcal{L}_{\text{AVD}}= \lambda\mathcal{L}_{\EE+\DD}+(1-\lambda)\mathcal{L}_{\EE+\TT}
    \end{aligned}
\end{equation}
where $\lambda$ is a regularization parameter. 

\section{Experiments}
\label{sec:experiment}

We evaluated the performance of our proposed AVD by conducting extensive experiments on three action recognition benchmarks, \ie UCF101~\cite{Soomro2012}, HMDB51~\cite{Kuehne2011}, and Kinetics~\cite{Kay2017}.

\subsection{Datasets}
\label{Sec:Datasets}

\noindent\textbf{UCF101:} The UCF101 dataset~\cite{Soomro2012} comprises realistic web videos, which are typically captured with large variations in camera motion, object appearance/scale, viewpoint, cluttered background, and illumination variations. It has $101$ categories of human actions ranging from daily life to sports. This dataset contains $13,320$ videos with an average duration of $7$ seconds. It comes with three split settings to separate the dataset into training and testing videos. We report the average classification accuracy over three splits.

\noindent\textbf{HMDB51:} The HMDB51 dataset~\cite{Kuehne2011} includes realistic videos from different sources, such as movies and web videos. The dataset has $6,766$ trimmed videos of $51$ human action classes. Similar to UCF101, this dataset has also three train/test splits and we report average classification accuracy of these three splits.

\noindent\textbf{Kinetics.} The Kintectics dataset~\cite{Kay2017} is a large-scale human action dataset, which consists of $400$ action classes where each category has more than $400$ videos. In total, there are about $240,000$ training videos, $20,000$ validation videos, and $40,000$ testing videos. The evaluation metrics on the Kinectics dataset is the average top-1 and top-5 accuracy. In our experiments, we only use RGB data of this dataset.

\subsection{Experimental Setup}
\label{sec:setup}

In our experiments, we used four deep architectures that have been pre-trained on ImageNet dataset~\cite{Russakovsky2015}. We  employed Tensorflow implementations of AlexNet~\cite{Krizhevsky2012}, Inception-V1~\cite{Ioffe2015}, ResNet-50~\cite{He2016}, and ResNet-101~\cite{He2016}. All of these deep models have been fine-tuned by using stochastic gradient descent with the momentum of $0.9$ and an annealed learning rate that is started with $3\times 10^{-3}$ and multiplied by a factor of $0.1$ per epoch. For the training, we randomly performed size jittering, cropping, flipping, and re-scaling on images. We also applyed our method on Optical Flow (OF) data. For the computation of the optical flow, we used TLV1 optical flow algorithm~\cite{Zach2007}, which is implemented in OpenCV with CUDA. The $\EE$ network consists of five layers of 3D convolutional kernels of size $5\times 5\times 3$ and strides 2, with batch normalization and ReLu layers added in between. The $\DD$ network mirrors the $\EE$ network, except that it uses Leaky ReLu instead of ReLu layers. The $\TT$ network comprises five fully connected layers, where we used ReLu between them and Sigmoid as the last layer. 

\subsection{Comparisons with Baseline Inputs}
\label{input_type}

 We evaluated the performance of AVD using different types of inputs, \ie RGB frames and Optical Flow (OF). RGB frames  and OF data are directly fed to the model. In this way, the information is extracted per frame and the average performance is reported. In addition, we applyed AVD on sequences of RGB frames and OFs data to achieve distilled representations. The ResNet-50~\cite{He2016} was used for video classification in this experiment. Table~\ref{tab:inputs} indicates the results of our experiments using different input types. By using RGB data, our AVD achieved $1.5\%$, $2.5\%$, $1.6\%$, and $1.7\%$ improvement over OF. We argue that this improvement is achieved due to exploiting appearance information along with motion information of RGB sequences. The AVD representation using RGB data still has superior performance compared to the baseline OF and RGB frames data.
\begin{table}[t]
\centering
\caption{The accuracy ($\%$) of our proposed method versus the accuracy of representations from optical flow and RGB data using ResNet-50~\cite{He2016}.}
\label{tab:inputs}
\resizebox{\linewidth}{!}{
\begin{tabular}{lcccc} \hline
 & \multirow{2}{*}{UCF101} & \multirow{2}{*}{HMDB51} & \multicolumn{2}{c}{Kinetics} \\
 &  &  & Top-1 & Top-5 \\ \hline \hline
RGB & 83.6 & 53.5 & 61.3 & 83.1 \\ 
OF & 86.5 & 58.0 & 64.7 & 87.5 \\ 
AVD (RGB) & 88.7 & 62.8 & 67.9 & 89.2 \\ 
AVD (OF) & 87.2 & 60.3 & 66.3 & 87.5 \\
OF+AVD (OF) & 91.1 & 64.0 & 68.5 & 89.2 \\
AVD (OF)+AVD (RGB) & 91.9 & 65.2 & 70.1 & 90.2 \\
OF+AVD (OF)+AVD (RGB) & \textbf{96.9} & \textbf{72.4} & \textbf{73.9} & \textbf{92.5} \\ \hline
\end{tabular}%
}
\end{table}

\subsection{AVD and Deep Models}
\label{deep_models}

We employed four deep models (namely AlexNet~\cite{Krizhevsky2012}, Inception-V1~\cite{Ioffe2015}, ResNet-50~\cite{He2016}, and ResNet-101~\cite{He2016}) to the representations obtained by the AVD to study the usefulness of various 2D CNN models in video classification. The results are reported in Table~\ref{tab:deep_models}. As expected, the best performances are achieved by deeper architectures. It is notwithstanding that given the significant reduction in the large amount of annotated videos for training a 3D model, AVD is an optimal alternative for enabling the usage of 2D CNNs in the video classification.
\begin{table}[t]
\centering
\caption{AVD performance evaluation ($\%$) using different 2D CNN models pre-trained on ImageNet dataset~\cite{Russakovsky2015}.}
\label{tab:deep_models}
\resizebox{\linewidth}{!}{%
\begin{tabular}{lcccc} \hline 
 & \multirow{2}{*}{UCF101} & \multirow{2}{*}{HMDB51} & \multicolumn{2}{c}{Kinetics} \\
 &  &  & Top-1 & Top-5 \\ \hline \hline
AlexNet~\cite{Krizhevsky2012} & 83.3 & 56.5 & 58.1 & 80.4 \\ 
Inception-V1~\cite{Ioffe2015} & 87.6 & 60.4 & 65.8 & 86.0 \\ 
ResNet-50~\cite{He2016} & 88.7 & 62.8 & 67.9 & 89.2 \\ 
ResNet-101~\cite{He2016} & 90.2 & 66.1 & 69.9 & 92.5 \\ \hline  
\end{tabular}%
}
\end{table}

\subsection{Cross-Dataset Analysis}
\label{sec:cross_dataset}

To gain insights into the generalization of distilled representations achieved by our AVD, we conducted cross-dataset experiments. In these experiments, we fine-tuned a pre-trained ResNet-50~\cite{He2016} using the distilled representations of training samples from one dataset and used the distilled representations of the test set of other datasets. Table~\ref{tab:crossdataset} summarizes the results. Despite the drop in the accuracy of the cross-dataset scenario, the AVD still shows a good performance, stating the high capability to represent video sequences discriminatively.
\begin{table}[t!]
\centering
\caption{The accuracy ($\%$) of our proposed AVD method in the cross-dataset experimental setting using ResNet-50~\cite{He2016}. The model is fine-tuned in one dataset and is tested on another dataset. We report the Top-1 accuracy for the Kinetics dataset.}
\label{tab:crossdataset}
\begin{tabular}{clccc}
\hline 
\multicolumn{1}{l}{}                                &                                  & \multicolumn{3}{c}{Test on} \\ \hline
\multicolumn{1}{l}{}                                &                                  & UCF101  & HMDB51  & Kinetics  \\ \hline
\multicolumn{1}{c}{\parbox[t]{2mm}{\multirow{3}{*}{\rotatebox[origin=c]{90}{Train on}}}} & \multicolumn{1}{l}{UCF101}      & \textbf{88.7}    & 59.7    & 60.1         \\ 
\multicolumn{1}{c}{}                               & \multicolumn{1}{l}{HMDB51}      & 83.3    & \textbf{62.8}    & 63.5         \\ 
\multicolumn{1}{c}{}                               & \multicolumn{1}{l}{Kinetics} & 85.9    & 61.0    & \textbf{67.9}         \\ \hline 
\end{tabular}
\end{table}

\subsection{Comparison against the State-of-the-Art}

We compared our AVD against the state-of-the-art methods. Table~\ref{tab:ucf_hmdb} summarizes the comparative results on UCF101~\cite{Soomro2012} and HMDB51~\cite{Kuehne2011} datasets. We compared our method to the both handcrafted methods
and deep learning based methods.
Among them,  Dynamic Image (DI)~\cite{Bilen2016} and SVM Pooled descriptor (SVMP)~\cite{Wang2018a} are conceptually the closest methods to ours. We represented the RGB videos using AVD. In our comparisons, we combined different baseline inputs for AVD. The obtained representations were fed to ResNet-50 and ResNet-101. DI have achieved $95.5\%$ and $72.5\%$ using a four stream network (still images, dynamic images, optical flow, and dynamic optical flow) and ResNext-101 on UCF101 and HMDB51, respectively.  AVD performed better than DI by increasing the classification accuracy by $1.8\%$ and $4.6\%$ using ResNet-101 on UCF101 and HMDB51, respectively. Note that ResNet-101 has a relatively lower power than ResNext-101 on images.
\begin{table}[t]
\small
\centering
\caption{Comparison of classification accuracy ($\%$) of the proposed approach against those of state-of-the-art methods on UCF101~\cite{Soomro2012} and HMDB51~\cite{Kuehne2011} datasets.}
\label{tab:ucf_hmdb}
\begin{tabular}{lcc}
\hline 
Method                    & UCF101        & HMDB51      \\ \hline \hline
DT+MVSV~\cite{Cai2014}                   & 83.5          & 55.9     \\     
iDT+HSV~\cite{Peng2016}                   & 87.9          & 61.1   \\
MoFAP~\cite{Wang2016}                     & 88.3          & 61.7    \\
Two-Stream~\cite{Simonyan2014}                & 88.0          & 59.4    \\
C3D (3 nets)~\cite{Tran2015}              & 85.2          & 51.6         \\
Res3D~\cite{Tran2017}                     & 95.6          & 54.9     \\
I3D~\cite{Carreira2017}                       & 95.6          & 74.8    \\
F\textsubscript{ST}CN~\cite{Sun2015}                     & 88.1          & 59.1        \\
LTC~\cite{Varol2018}                       & 91.7          & 64.8     \\  
KVMF~\cite{Zhu2016}                      & 93.1          & 63.3       \\
TSN (7 seg)~\cite{Wang2018}               & 94.9          & 71.0  \\   
DI (4 stream)~\cite{Bilen2016}             & 95.5          & 72.5          \\
SVMP~\cite{Wang2018a}                      & 94.6             & 71.0   \\   
S3D-G~\cite{Xie2018}                     & 96.8          & 75.9          \\
     \hline
\textbf{AVD (ResNet-50)}  & \textbf{96.9} & \textbf{72.4}  \\
\textbf{AVD (ResNet-101)} & \textbf{97.3} & \textbf{77.1} \\
 \hline 
\end{tabular}
\end{table}

We also compared the performance of our method using Kinetics dataset, which is a very large dasaset. In Table~\ref{tab:kinetics}, we reported the results of the comparison of our AVD with the state-of-the-art methods in terms of Top-1 and Top-5 accuracy. As can be seen, the AVD achieves $75.1\%$ and $93.5\%$ Top-1 and Top-5 accuracy, respectively. The results of this experiment show that our AVD can handle the inter-class similarity of different action categories and still achieves discriminative representations for different classes.
\begin{table}[t!]
\centering
\caption{Comparison of our proposed method's performance in terms of Top-1 and Top-5 classification accuracy ($\%$) against the state-of-the-art methods on the Kinetics dataset~\cite{Kay2017}.}
\label{tab:kinetics}
\begin{tabular}{lcc}
\hline 
Method              & Top-1         & Top-5    \\ \hline \hline
Two-Stream~\cite{Simonyan2014}          & 61.0          & 83.3     \\    
Two-Stream (I3D)~\cite{Carreira2017}    & 74.9          & 91.8  \\
I3D (RGB)~\cite{Carreira2017}           & 72.9          & 90.8          \\
I3D (OF)~\cite{Carreira2017}            & 63.9          & 85.0      \\    
I3D (RGB+OF)~\cite{Wang2018}        & 74.1          & 91.6 \\
TSN~\cite{Wang2018}                 & 73.9          & 91.1       \\  
TSN (I3D)~\cite{Wang2018}           & 75.7          & 92.5      \\
S3D-G (RGB)~\cite{Xie2018}         & 74.7          & 93.4      \\  
S3D-G (RGB+OF)~\cite{Xie2018}      & 77.2          & 93.0    \\
    \hline
\textbf{AVD (ResNet-50)}          & 73.9          & 92.5    \\     
\textbf{AVD (ResNet-101)} & \textbf{75.1} & \textbf{93.4} \\
  \hline 
\end{tabular}
\end{table}

\section{Conclusion}
\label{sec:conclusion}

In this paper, we presented an adversarial video distillation for capturing and encoding the appearance and motion of the video into one single image, which can be processed by deep models pre-trained on still images. This technique tackled the problem of tuning huge number of parameters in deep models for videos. We extended an adversarial autoencoder by adding a generative model, which serves as a decoder network to a 3D convolutional encoder network. We trained the entire model in an unsupervised adversarial end-to-end manner. The encoder network learned to extract semantic representations from a given input video by mapping the 3D input into a 2D latent representation. Once the model was trained, we used 2D representations obtained by the encoder network as the input to pre-trained deep models for video classification. This distilled image representation includes the gist of the video. We evaluated the effectiveness of the proposed method on three benchmark datasets, namely, UCF101, HMDB51, and Kinetics, for video representation and classification. The experimental results demonstrated the superior performance of our proposed method.

{\small
\bibliographystyle{ieee_fullname}

}

\end{document}